\documentclass[conference]{IEEEtran}
\IEEEoverridecommandlockouts

\usepackage{amsmath,amssymb,bm}
\usepackage{siunitx}
\usepackage{graphicx}
\usepackage{booktabs}
\usepackage{xcolor}
\usepackage{float}
\usepackage{algorithm}
\usepackage{algpseudocode}
\usepackage{amsthm}
\usepackage{multirow}
\usepackage[colorlinks=true,linkcolor=black,citecolor=blue,urlcolor=black]{hyperref}
\usepackage[capitalize,noabbrev]{cleveref}
\crefname{equation}{Eq.}{Eqs.}
\Crefname{equation}{Equation}{Equations}
\usepackage[nopatch=footnote]{microtype}
\usepackage{placeins}
\usepackage{setspace}
\newcommand{\vect}[1]{\bm{#1}}

\title{\LARGE \bf
Decoupling Torque and Stiffness: A Unified Modeling and Control Framework for Antagonistic Artificial Muscles
}
\author{Amirhossein Kazemipour$^{1}$ and Robert K. Katzschmann$^{1,*}$
\thanks{$^{1}$Soft Robotics Lab, IRIS, D-MAVT, ETH Zurich, Switzerland
{\texttt{\{akazemi, rkk\}@ethz.ch}}}%
\thanks{$*$ Corresponding author: \href{mailto:rkk@ethz.ch}{\tt rkk@ethz.ch}}
}

\begin{document}
\raggedbottom
\maketitle

\begin{abstract}
Antagonistic artificial muscles can decouple joint torque and stiffness, but contact transients often degrade this independence. We present a unified real-time framework applicable across pneumatic, electrohydraulic, and dielectric elastomer artificial muscle families: a separable Pad\'e force model with a minimal two-state dynamic wrapper, a cascaded inverse-dynamics controller in co-contraction/bias coordinates, and a bio-inspired depth-adaptive interaction policy that schedules stiffness based on penetration depth. The controller runs in under \SI{1}{ms} per control tick and demonstrates independent torque and stiffness tracking, including a fixed-torque stiffness-step test that preserves torque regulation through stiffness transitions. In a coupled impedance contact protocol simulated across soft-to-rigid environments, comparing depth-adaptive stiffness to fixed-stiffness baselines reveals a shock/load versus stability tradeoff. These results provide a control-oriented foundation for musculoskeletal antagonistic robots to execute adaptive impedance behaviors in dynamic interactions.
\end{abstract}

\begin{IEEEkeywords}
Antagonistic Actuation, Artificial Muscles, Variable Impedance Control, Dynamic Modeling, System Identification.
\end{IEEEkeywords}

\section{Introduction}
Animals regulate joint behavior with antagonistic muscle pairs, using \emph{bias} to produce net torque and \emph{co-contraction} to tune mechanical stiffness \cite{Zajac1989}. During unexpected contact, intrinsic muscle properties and fast reflex pathways can adjust effective stiffness on short timescales \cite{houk1981regulation}, enabling robust interactions during rapid contact events \cite{burdet2001optimalimpedance}. This motivates musculoskeletal robots that can command torque and stiffness as independent physical quantities, especially during contact transients.

\begin{figure}[!t]
    \centering
    \includegraphics[width=0.98\columnwidth,clip,trim=2 2 2 0]{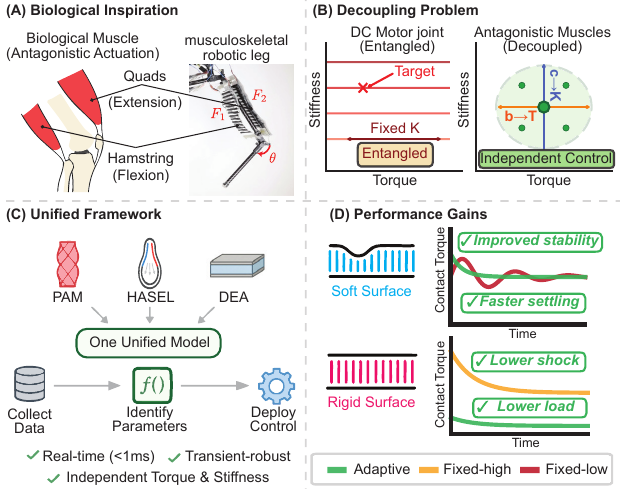}
    \caption{\textbf{(A)} Biological inspiration: antagonistic muscle pairs support independent torque-stiffness control via bias and co-contraction. \textbf{(B)} In antagonistic artificial muscles, dual activations enable plant-level stiffness commandability; in co-contraction/bias coordinates $c_{\alpha}=(\alpha_1+\alpha_2)/2$ and $b_{\alpha}=(\alpha_1-\alpha_2)/2$, $c_{\alpha}$ modulates stiffness $K$ and $b_{\alpha}$ generates torque $T$. \textbf{(C)} Unified framework: a separable Pad\'e [2/1] force model with a minimal dynamic wrapper provides a common model form applicable across major artificial-muscle families for sub-millisecond real-time control. \textbf{(D)} Simulation: a depth-adaptive stiffness schedule reveals a shock/load versus stability tradeoff across soft-to-rigid contacts and produces responses that interpolate between low and high fixed-stiffness baselines across contact regimes.}
    \label{fig:introduction}
\end{figure}

\begin{figure*}[t]
    \centering
    \includegraphics[width=0.985\textwidth,clip,trim=1 1 1 1]{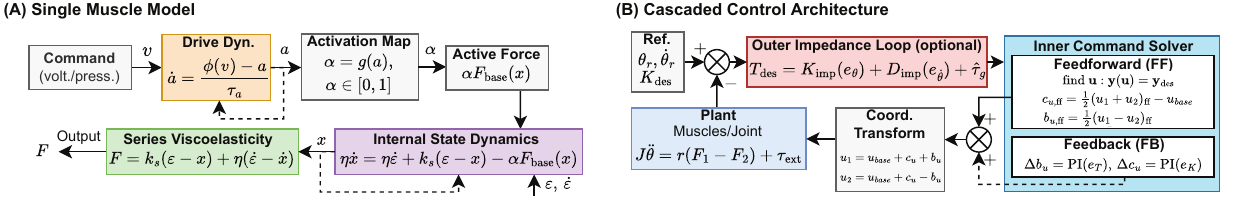}
    \caption{Overview of the unified muscle model and cascaded controller. \textbf{(A)} A two-state muscle model combines activation dynamics, a separable Pad\'e base law, and a viscoelastic series element. \textbf{(B)} A cascaded controller supports two operating modes: direct $(T_{\text{des}},K_{\text{mech,des}})$ commands for plant-level decoupling tests, and an optional outer impedance loop for interaction tasks, both implemented via a command-space inner solver and lightweight feedback in co-contraction/bias coordinates around the preload command level $u_{\text{base}}$.}
    \label{fig:overview}
\end{figure*}

Soft artificial muscles, including pneumatic artificial muscles (PAMs, e.g., McKibben muscles), hydraulically amplified self-healing electrostatic actuators (HASELs), and dielectric elastomer actuators (DEAs), are attractive building blocks for such systems because they combine compliance and power density \cite{Tondu2012,Kellaris2018,acome2018hydraulically}. They are increasingly used in musculoskeletal robots \cite{kurumaya2016musculoskeletal,buchner2024electrohydraulic,kazemipour2025stretchable}. However, maintaining independent torque and stiffness control in antagonistic pairs remains challenging in practice: actuator force depends nonlinearly on strain and input, and rate limits and viscoelastic effects are most pronounced during the first tens of milliseconds after contact.

In this work, we present and evaluate a framework for \emph{independent regulation of joint torque and mechanical stiffness that remains robust during contact transients}. Concretely, we evaluate three capabilities: (i) real-time tracking of independent torque and stiffness commands, (ii) torque regulation through rapid stiffness transitions (fixed-torque stiffness-step test), and (iii) a single depth-adaptive policy that captures the impact-mitigation versus stability tradeoff across soft-to-rigid contacts without explicit surface labels. These results move beyond nominal decoupling by targeting behavior that remains reliable and interpretable when contact dynamics change abruptly.

To achieve these capabilities, we combine a unified low-order muscle model with a cascaded inverse-dynamics controller that exposes a direct $(T_{\text{des}},K_{\text{mech,des}})$ interface, and we evaluate it in early-contact interaction tests with attribution ablations.

Prior work provides key foundations: impedance control in rigid robotics \cite{hogan1985impedance}, plant-level stiffness modulation in VSA/VIA systems \cite{Vanderborght2013,Albu-Schaeffer2008}, and decoupled position and stiffness control on specific antagonistic soft platforms \cite{trumic2020decoupled,best2020comparing,habich2023learning}. Our contribution is not a first demonstration of decoupling. Instead, we focus on the early-contact regime, where preserving the three capabilities above is difficult and where it is often unclear whether observed behavior is driven by outer-loop impedance scheduling or by plant-level stiffness modulation.

This regime is particularly challenging for two reasons: quasi-static models can be insufficient immediately after impact, and actuator-specific high-fidelity simulators are often expensive to invert in real time \cite{Tondu2012,Chou1996,gravert2024low,wissler2007electromechanical,liu2023optimization}. We address this with a unified, control-oriented dynamic muscle framework consisting of a separable Pad\'e base law, a minimal two-state wrapper, and a cascaded inverse-dynamics pipeline. We evaluate the framework under one reproducible contact protocol and include explicit attribution ablations that bound outer-loop impedance scheduling versus plant-level stiffness modulation during fast transients. To keep comparisons fair and interpretable, the contact study uses paired randomized perturbations shared across policies.

Figures~\ref{fig:introduction} and~\ref{fig:overview} summarize the high-level concept and model/controller operating modes used throughout Sections~II--V. We use $K_{\text{mech}}$ for inner-loop mechanical stiffness, $K_{\text{tot}}$ for total joint stiffness, and $K_{\text{imp}}$ for outer-loop virtual gains.

Specifically, this paper contributes three elements: (1) a separable Pad\'e [2/1] force model with actuator-specific activation mappings, augmented with a minimal two-state dynamic wrapper for bandwidth and viscoelastic effects; (2) a cascaded inverse-dynamics controller that enables independent torque and stiffness tracking in under \SI{1}{ms} per control tick; and (3) a simulation-based contact evaluation, including a fixed-torque stiffness-step capability test and depth-adaptive interaction studies with attribution ablations that bound the role of outer-loop impedance scheduling versus plant-level stiffness modulation.

\section{Unified Modeling Framework}

\subsection{Separable Pad\'e Base Force Law}
We model each muscle with a separable active force and use the same formulation across PAMs/McKibben muscles, HASELs, and DEAs; here we validate on HASEL hardware, while cross-actuator hardware validation is future work.
\begin{equation}
    F(\alpha,x) = \alpha\,F_{\text{base}}(x), \qquad F_{\text{base}}(x) = \frac{c_0 + c_1 x + c_2 x^2}{1 + d_1 x},
    \label{eq:pade}
\end{equation}
where $\alpha\in[0,1]$ is normalized activation and $x$ is the internal deformation state of the active element (defined in Section~II-B). The Padé [2/1] form uses four coefficients and retains low complexity for real-time control while capturing nonlinear force-length behavior. The coefficient vector is $\vect{p} = (c_0,c_1,c_2,d_1)$, with small-deformation slope $k_a = c_1 - c_0 d_1$. For physical validity, we require $F_{\text{base}}(x)\ge 0$ and $1+d_1 x>0$ over the operating interval. The derivative of the Pad\'e base force law is
\begin{equation}
    F'_{\text{base}}(x)=\frac{(c_1-d_1c_0)+2c_2 x + d_1 c_2 x^2}{(1+d_1 x)^2},
    \label{eq:pade_deriv}
\end{equation}
which will be used for stiffness calculations.

This Pad\'e [2/1] formulation can represent diverse muscle physics with actuator-specific activation mappings:

\textbf{PAMs/McKibben:} From the geometric model \cite{Chou1996}, the force expands to $F = P \cdot (q_0 + q_1\varepsilon + q_2\varepsilon^2)$ after strain conversion. Setting $\alpha=P/P_{\max}$ and $d_1=0$ in our framework recovers the common quadratic approximation used for PAM force--strain curves, with $c_i = P_{\max} \cdot q_i$.

\textbf{HASELs:} Maxwell stress yields force that depends nonlinearly on strain \cite{Kellaris2018}. Using $\alpha=(V/V_{\max})^2$, the Padé form with $d_1 \neq 0$ captures the characteristic force roll-off at high strains. The denominator term models electrohydraulic coupling where hydraulic constraints limit deformation.

\textbf{DEAs:} For DEAs, force can be linearized around an operating point in simplified regimes \cite{wissler2007electromechanical}. With $\alpha=(V/V_{\max})^2$ (equivalently $(E/E_{\max})^2$ under fixed geometry), setting $d_1=0$ and $c_2=0$ yields $F = \alpha(c_0 + c_1x)$ (and in quasi-static conditions $x=\varepsilon$).

The coefficient $d_1$ distinguishes actuator complexity: $d_1=0$ for polynomial force-strain relations (PAMs, simple DEAs), while $d_1 \neq 0$ captures saturation-like actuator-specific nonlinearities (e.g., HASELs with hydraulic limits).
In the HASEL implementation used here, $\alpha=(V/V_{\max})^2$ is precomputed from voltage data, and runtime commands operate in normalized activation coordinates.

During identification, we impose mild shape constraints for physical plausibility: positivity of $F_{\text{base}}$, monotonicity over the operating contraction range, denominator regularity, and low force near the maximum contraction bound used in Section~II-C.

\subsection{Minimal Two-State Dynamic Wrapper}
Finite input bandwidth and internal viscoelasticity are captured with two actuator-side states; $\varepsilon(t)$ denotes external contraction strain from joint geometry, while $x(t)$ is the internal active-element deformation used in the Pad\'e law. A drive state $a(t)$ filters the physical actuator command $v(t)$:
\begin{equation}
    \dot a = \frac{\phi(v) - a}{\tau_a}, \qquad \alpha = g(a),
    \label{eq:drive}
\end{equation}
where $v$ is the physical actuator command (e.g., voltage or pressure), $\phi$ maps actuator command space to a drive level, and $g$ maps drive state to realized activation $\alpha \in [0,1]$.

A Kelvin--Voigt branch in series with the active element introduces an internal deformation $x(t)$ such that
\begin{align}
    F &= k_s (\varepsilon - x) + \eta (\dot\varepsilon - \dot x), \label{eq:series_branch}\\
    \eta \dot x &= \eta \dot\varepsilon + k_s (\varepsilon - x) - \alpha F_{\text{base}}(x).
    \label{eq:x_dynamics}
\end{align}
The parameters $k_s \ge 0$ and $\eta\ge 0$ represent series stiffness and damping, respectively. When $\tau_a \rightarrow 0$, $k_s \rightarrow \infty$, and $\eta \rightarrow 0$ the quasi-static force law is recovered.

\subsection{Antagonistic Joint with Gravity}
For a revolute joint with moment arm $r$ and reference length $L_0$, we use explicit installation-slack geometry and define $\xi \triangleq r/L_0$. Let $S$ denote the tendon take-up displacement at $\theta=0$. The required tendon displacements are
\begin{equation}
    x_{\mathrm{req},1}=S+r\theta,\qquad x_{\mathrm{req},2}=S-r\theta,
\end{equation}
and the physical contraction strains passed to the Pad\'e law are
\begin{equation}
    \varepsilon_i=\mathrm{clip}\!\left(\frac{x_{\mathrm{req},i}}{L_0},\,0,\,\varepsilon_{\max}\right),\quad i\in\{1,2\}.
\end{equation}
Joint dynamics with equivalent inertia $J_{\mathrm{eq}}$, passive elements $(B_j, K_j)$, and gravity torque $\tau_g(\theta) = m_\ell g \ell_c \sin(\theta - \theta_g)$ (where $\theta_g$ is the gravity-neutral angle, set to 0 in our simulations) become
\begin{equation}
    J_{\mathrm{eq}} \ddot\theta + B_j \dot\theta + K_j \theta + \tau_g(\theta) = r (F_1 - F_2) + \tau_{\mathrm{ext}}.
    \label{eq:eom}
\end{equation}
Here $F_i$ denotes the total force produced by antagonistic side $i$ (including all parallel packs, if present).
The corresponding posture-dependent gravity torque slope is
\begin{equation}
    K_g(\theta) \triangleq \frac{\partial \tau_g}{\partial \theta} = m_\ell g \ell_c \cos(\theta - \theta_g).
\end{equation}
We define torque slopes as $K \triangleq \partial \tau/\partial \theta$. When these terms are written as restoring components on the left-hand side, we report their positive magnitudes.

\paragraph{Slack handling} Tendons can only pull ($F_i \ge 0$). We model this explicitly through $(x_{\mathrm{req},1},x_{\mathrm{req},2})$ and enforce non-stretchability with geometric bounds $0\le x_{\mathrm{req},i}\le x_{\max}$, where $x_{\max}=\varepsilon_{\max}L_0$. The resulting feasible interval
\begin{equation}
    \begin{aligned}
        \theta_{\min}&=\max\!\left(-\frac{S}{r},\frac{S-x_{\max}}{r}\right),\\
        \theta_{\max}&=\min\!\left(\frac{S}{r},\frac{x_{\max}-S}{r}\right).
    \end{aligned}
\end{equation}
is enforced in simulation by hard projection of $\theta$. The control offset $u_{\text{base}}$ sets the symmetric pre-activation operating point; optional angle-dependent command floors are used only in dedicated audit runs.

\subsection{Stiffness Derivation}
The per-muscle restoring stiffness uses a positive-magnitude convention: $k_{\mathrm{act},i}\triangleq\alpha_i\,|F'_{\text{base}}(x_i)|$ (with $F'_{\text{base}}$ from \cref{eq:pade_deriv}), combined in series with the Kelvin--Voigt branch. For discrete-time perturbations at interval $\Delta t$ (set to the controller update period, \SI{1}{ms}), the damping contributes an effective stiffness $\eta/\Delta t$, giving series stiffness $K_s = k_s + \eta/\Delta t$. The harmonic combination of active and series elements yields:
\begin{equation}
    K_{i,\text{step}} = \frac{k_{\mathrm{act},i}\,K_s}{k_{\mathrm{act},i}+K_s}.
    \label{eq:Kistep}
\end{equation}
This is the magnitude-based series combination used in the implementation; in our regime $K_s \gg k_{\mathrm{act},i}$, so it closely matches the exact one-step linearization. The joint-level mechanical stiffness contribution is $K_{\text{mech,step}} = r \xi \big(K_{1,\text{step}} + K_{2,\text{step}}\big)$ with $\xi=r/L_0$. The total joint stiffness slope is
\begin{equation}
    K_{\text{tot,step}} = K_{\text{mech,step}} + K_j + K_g(\theta).
    \label{eq:Ktot}
\end{equation}
For interpretation, we report $K_{\text{mech,step}}$ and $K_{\text{tot,step}}$ as positive restoring magnitudes in the linearized left-hand-side stiffness term; equivalently, for the actuator torque on the right-hand side, $K_{\text{mech,step}} \approx -\partial\!\left[r(F_1-F_2)\right]/\partial\theta$ near the operating point. This defines the conversion between total and mechanical stiffness used in the cascaded architecture. For brevity in the control sections, we drop the ``step'' subscript and use $K_{\text{mech}}$ and $K_{\text{tot}}$ to denote the discrete-time step stiffness from \cref{eq:Kistep,eq:Ktot} with $\Delta t=\SI{1}{ms}$. The antagonistic joint admits a natural coordinate transformation for decoupling:
\begin{equation}
    c_{\alpha} \;\triangleq\; \tfrac{1}{2}(\alpha_1+\alpha_2), \qquad b_{\alpha} \;\triangleq\; \tfrac{1}{2}(\alpha_1-\alpha_2),
    \label{eq:co_contraction_bias}
\end{equation}

\section{Model Identification and Validation}
\subsection{Experimental Model Identification}
We validate the muscle modeling framework using a custom-designed muscle characterization testbed that records force, position, and velocity for commanded voltages to HASEL actuators at \SI{50}{\hertz} (sufficient for the identified bandwidth in this study; identifying faster transients would require higher-rate sensing). The identification procedure generalizes to any artificial muscle via appropriate activation mappings ($\alpha=P/P_{\max}$ for PAMs, $\alpha=(V/V_{\max})^2$ for HASELs, and $\alpha=(V/V_{\max})^2$ for DEAs, equivalent to $(E/E_{\max})^2$ under fixed geometry). The identification exploits time-scale separation: quasi-static force-strain data from slow, high-activation trials fit the Pad\'e base curve (\cref{eq:pade}) under physics constraints (non-negativity, monotonicity, denominator regularity), while the full dynamic dataset identifies series parameters $(k_s, \eta, \tau_a)$ and validates transient response. All identified muscle parameters and simulation joint parameters are provided in Table~\ref{tab:params}.

\begin{table}[!t]
    \caption{Model parameters from experimental identification (muscle) and simulation setup (joint).}
    \label{tab:params}
    \centering
    \small
    \begin{tabular}{@{}lrl@{\hspace{1.5em}}lrl@{}}
        \toprule
        \multicolumn{3}{c}{\textbf{Muscle Parameters (Learned)}} & \multicolumn{3}{c}{\textbf{Joint \& Link Parameters}} \\
        \cmidrule(r){1-3} \cmidrule(l){4-6}
        Param. & Value & Unit & Param. & Value & Unit \\
        \midrule
        $c_0$ & \num{6.804} & N & $r$ & \num{0.02} & m \\
        $c_1$ & \num{-171.076} & N & $L_0$ & \num{0.16} & m \\
        $c_2$ & \num{1087.818} & N & $J_{\mathrm{eq}}$ & \num{0.31} & kg$\cdot$m$^2$ \\
        $d_1$ & \num{5.674} & -- & $B_j$ & \num{1.0} & \si{\newton\meter\second\per\radian} \\
        $k_s$ & \num{2370.9} & N & $K_j$ & \num{1.0} & \si{\newton\meter\per\radian} \\
        $\eta$ & \num{64.98} & N$\cdot$s & $m_\ell$ & \num{1.0} & kg \\
        $\tau_a$ & \num{0.040} & s & $\ell_c$ & \num{0.1} & m \\
         &  &  & $g$ & \num{9.81} & m/s$^2$ \\
         &  &  & $\theta_g$ & \num{0.0} & rad \\
         &  &  & $S$ & \num{0.004} & m \\
         &  &  & $\varepsilon_{\max}$ & \num{0.05} & -- \\
         &  &  & $n_{\text{packs}}$ & \num{60} & -- \\
        \bottomrule
    \end{tabular}
\end{table}

\begin{figure}[!tb]
    \centering
    \includegraphics[width=0.95\columnwidth]{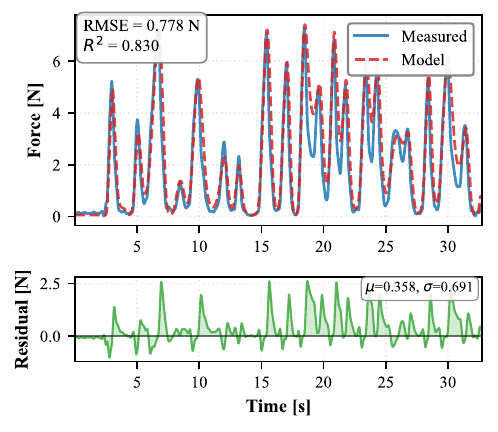}
    \caption{Held-out force prediction on an experimental trajectory. The model (red, dashed) matches measurements (blue) with RMSE = \SI{0.78}{N} and $R^2=0.83$, capturing both quasi-static trends and transient response.}
    \label{fig:identification}
\end{figure}

The model achieves sub-Newton accuracy on held-out hardware data (RMSE = \SI{0.78}{N}, $R^2{=}0.83$), confirming that the same low-order structure with actuator-specific activation mapping captures measured HASEL force dynamics without actuator-type-specific states. In the measured \SIrange{7}{8}{N} range this is about 10\% relative error. This is acceptable here because residuals are corrected by closed-loop torque/stiffness feedback and contact claims rely on paired policy ordering. Residuals are largest during sharp transients, motivating future higher-order viscoelastic/hysteresis extensions.

\subsection{Simulation Validation}
\begin{figure}[!t]
    \centering
    \includegraphics[width=0.83\linewidth,clip,trim=3 2 3 0]{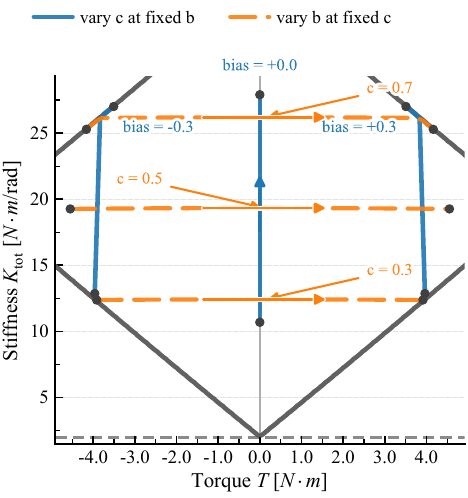}
    \caption{Torque-stiffness decoupling at $\theta=0^\circ$. Sweeping co-contraction $c_{\alpha}$ at fixed bias $b_{\alpha}$ (blue) changes total joint stiffness $K_{\text{tot}}$ with minimal torque change, while sweeping $b_{\alpha}$ at fixed $c_{\alpha}$ (orange) changes torque with minimal stiffness change. Gray indicates the feasible region; the horizontal axis denotes joint torque $T$.}
    \label{fig:decoupling-panel}
\end{figure}

Numerical integration uses \SI{0.5}{\milli\second} timesteps with parameters from Table~\ref{tab:params}.
Figure~\ref{fig:decoupling-panel} demonstrates the separation structure: varying co-contraction $c_{\alpha}=\tfrac{1}{2}(\alpha_1+\alpha_2)$ at fixed bias modulates stiffness with negligible torque change (blue, vertical), while varying bias $b_{\alpha}=\tfrac{1}{2}(\alpha_1-\alpha_2)$ at fixed co-contraction produces torque swings with minimal stiffness variation (orange, horizontal). The gray boundary shows actuation limits. This validates the torque-stiffness decoupling predicted by the stiffness formulation in \cref{eq:Kistep,eq:Ktot}.

\section{Model-Based Control Synthesis}
\subsection{Cascaded Architecture}
We present a cascaded controller comprising an outer impedance loop and an inner command solver. The full cascaded system operates as follows:

\paragraph{Outer-loop impedance law} Given position references $(\theta_r,\dot\theta_r)$ and desired impedance parameters $(K_{\text{imp}}, D_{\text{imp}})$, the outer-loop computes the desired torque
\begin{equation}
    T_{\text{des}} = K_{\text{imp}} (\theta_r - \theta) + D_{\text{imp}} (\dot\theta_r - \dot\theta) + \hat{\tau}_g(\theta),
    \label{eq:outer_loop}
\end{equation}
where $\hat{\tau}_g$ compensates gravity, and $K_{\text{imp}}$, $D_{\text{imp}}$ are the impedance gains defining how strongly torque responds to tracking errors. This generates torque commands from position errors for compliant trajectory tracking tasks.

\paragraph{Inner-loop command solver} The inner-loop solves for normalized commands $\vect{u}=[u_1\ u_2]^\top$ to realize the targets $(T_{\text{des}},K_{\text{mech,des}})$ while respecting \cref{eq:drive,eq:series_branch,eq:x_dynamics,eq:Ktot}. For actuator $i$, the corresponding physical command entering \cref{eq:drive} is
\begin{equation}
    v_i = v_{i,\min} + (v_{i,\max}-v_{i,\min})\,u_i, \qquad i\in\{1,2\},
    \label{eq:u_to_v}
\end{equation}
with $u_i \in [u_{\min},u_{\max}] \subseteq [0,1]$. This reduces to $v_i=v_{i,\max}u_i$ when $v_{i,\min}=0$ (e.g., voltage-driven HASELs in this study). Here $K_{\text{mech,des}}$ is the mechanical stiffness contribution tracked by the inner-loop. The scalar $u_{\text{base}}$ denotes the common preload command level, equal drive to both muscles that sets the nominal pretension operating point used by the inner-loop. We assemble muscle torque and stiffness into
\begin{equation}
    \vect{y}(\vect{u}) = \begin{bmatrix}T(\vect{u})\\ K_{\text{mech,step}}(\vect{u})\end{bmatrix}
    = \begin{bmatrix}r\big(\alpha_1 F_{\text{base}}(x_1) - \alpha_2 F_{\text{base}}(x_2)\big)\\ r\xi\big(K_{1,\text{step}} + K_{2,\text{step}}\big) \end{bmatrix},
    \label{eq:inner_outputs}
\end{equation}
with $K_{i,\text{step}}$ from \cref{eq:Kistep}. In implementation, $\vect{y}(\vect{u})$ is the one-step prediction $\vect{y}_{k+1|k}(\vect{u})$ obtained by propagating \cref{eq:drive,eq:series_branch,eq:x_dynamics} over one control tick ($\Delta t=\SI{1}{ms}$) from the current state. The target is $\vect{y}_{\text{des}}=[T_{\text{des}},\ K_{\text{mech,des}}]^\top$, so the error is $\vect{e}(\vect{u}) = \vect{y}(\vect{u})-\vect{y}_{\text{des}}$. If a total-stiffness target is specified, we convert it via $K_{\text{mech,des}}=K_{\text{tot,des}}-K_j-K_g(\theta)$. Activations $\alpha_i$ follow from \cref{eq:drive} after mapping $\vect{u}\mapsto\vect{v}$ with \cref{eq:u_to_v}.

\paragraph{Operating modes and plant-level decoupling} The cascaded architecture supports two distinct operating modes: (1)~\textbf{Direct torque-stiffness control}, where $(T_{\text{des}}, K_{\text{mech,des}})$ are externally specified based on task requirements, demonstrating pure plant-level decoupling (Section~\ref{sec:controller_comparison}); and (2)~\textbf{Impedance control mode}, where the outer-loop generates $T_{\text{des}}$ from position errors via \cref{eq:outer_loop}, implementing policy-level coupling suitable for compliant interaction tasks. The contact studies in Section~\ref{sec:contact} use both modes: the fixed-torque stiffness-step capability test uses mode~(1), while the coupled policy audit uses mode~(2), where the outer impedance law generates $T_{\text{des}}$ and policy differences are expressed through $K_{\text{mech,des}}$ scheduling. Decoupling remains essential because the inner-loop must still track torque and stiffness targets through independent actuator channels despite contact transients and model uncertainty. Single-channel motor plants render stiffness in software, whereas VSA/VIA systems (including antagonistic architectures) can modulate plant stiffness; here we target antagonistic muscles because, in the studied architecture, they realize this with two muscle activations and no dedicated extra stiffness actuator.

\subsection{Feedforward Open-Loop Controller}
The feedforward controller finds normalized commands $\vect{u}^*$ satisfying $\vect{y}(\vect{u}^*) = \vect{y}_{\text{des}}$ via damped Gauss-Newton (Levenberg-Marquardt) iteration with the Jacobian
\begin{equation}
    J_u(\vect{u}) = \frac{\partial \vect{y}}{\partial \vect{u}}
    \approx \begin{bmatrix}\partial T/\partial u_1 & \partial T/\partial u_2\\ \partial K_{\text{mech,step}}/\partial u_1 & \partial K_{\text{mech,step}}/\partial u_2\end{bmatrix}.
    \label{eq:inner_jac}
\end{equation}
In implementation, $J_u$ is computed numerically by symmetric finite differences on $\vect{y}(\vect{u})$ with a normalized probe step $\Delta u_{\text{probe}}=0.01$, and each tick is warm-started from the previous command. We then update $\Delta \vect{u} = -(J_u^\top J_u + \lambda I)^{-1} J_u^\top \vect{e}$ with a regularization $\lambda = 10^{-4}$. The solution converges in 1 to 2 iterations when warm-started, with final normalized commands $u_{i,\text{cmd}}=\Pi_{[u_{\min},u_{\max}]}(u_i+\Delta u_i)$ and command-space slew limits.

\subsection{Closed-Loop Controller}
The closed-loop controller augments feedforward with PI feedback on errors $e_T = T_{\text{des}} - \hat{T}$ and $e_{K_{\text{mech}}} = K_{\text{mech,des}} - \hat{K}_{\text{mech}}$. In simulation, $\hat{T}$ and $\hat{K}_{\text{mech}}$ are the previous-tick plant outputs (with optional injected delay/noise in challenge profiles). We implement feedback in normalized command coordinates defined on offset-removed commands $\tilde u_i \triangleq u_i-u_{\text{base}}$:
\begin{equation}
    \begin{aligned}
        c_u &\;\triangleq\; \tfrac{1}{2}(\tilde u_1+\tilde u_2)=\tfrac{1}{2}(u_1+u_2)-u_{\text{base}},\\
        b_u &\;\triangleq\; \tfrac{1}{2}(\tilde u_1-\tilde u_2)=\tfrac{1}{2}(u_1-u_2),
    \end{aligned}
    \label{eq:command_coords}
\end{equation}
which preserve the co-contraction/bias semantics of \cref{eq:co_contraction_bias} because the mappings $u_i \mapsto v_i$ (\cref{eq:u_to_v}) and $v_i \mapsto \alpha_i$ (\cref{eq:drive}) are monotone over the operating range. The PI updates are
\begin{equation}
    \begin{aligned}
        \Delta b_u &= K_{pT} e_T + K_{iT} \int e_T dt,\\
        \Delta c_u &= K_{pK} e_{K_{\text{mech}}} + K_{iK} \int e_{K_{\text{mech}}} dt
    \end{aligned}
    \label{eq:pi_combined}
\end{equation}
Combining with feedforward $(c_{u,\text{ff}}, b_{u,\text{ff}})$ yields
\begin{equation}
    c_{u,\text{cmd}} = c_{u,\text{ff}} + \Delta c_u, \qquad b_{u,\text{cmd}} = b_{u,\text{ff}} + \Delta b_u,
    \label{eq:pi_cmd_coords}
\end{equation}
and normalized muscle commands
\begin{equation}
    \begin{aligned}
        u_{1,\text{cmd}} &= \Pi\!\left(u_{\text{base}}+c_{u,\text{cmd}}+b_{u,\text{cmd}}\right),\\
        u_{2,\text{cmd}} &= \Pi\!\left(u_{\text{base}}+c_{u,\text{cmd}}-b_{u,\text{cmd}}\right),
    \end{aligned}
    \label{eq:u_cmd_map}
\end{equation}
with anti-windup protection and command-space rate limiting; physical actuator commands are then obtained via \cref{eq:u_to_v}. The offset $u_{\text{base}}$ sets the symmetric command operating point (equal command to both muscles), selected from a neutral target stiffness in our experiments.

PI gains are designed from local plant sensitivities with torque bandwidth \SI{4.0}{\hertz}, stiffness bandwidth \SI{1.5}{\hertz}, and damping ratio $\zeta = 1.0$, ensuring stability while rejecting disturbances.

\subsection{Controller Comparison}\label{sec:controller_comparison}
\begin{figure}[htbp]
    \centering
    \includegraphics[width=\linewidth]{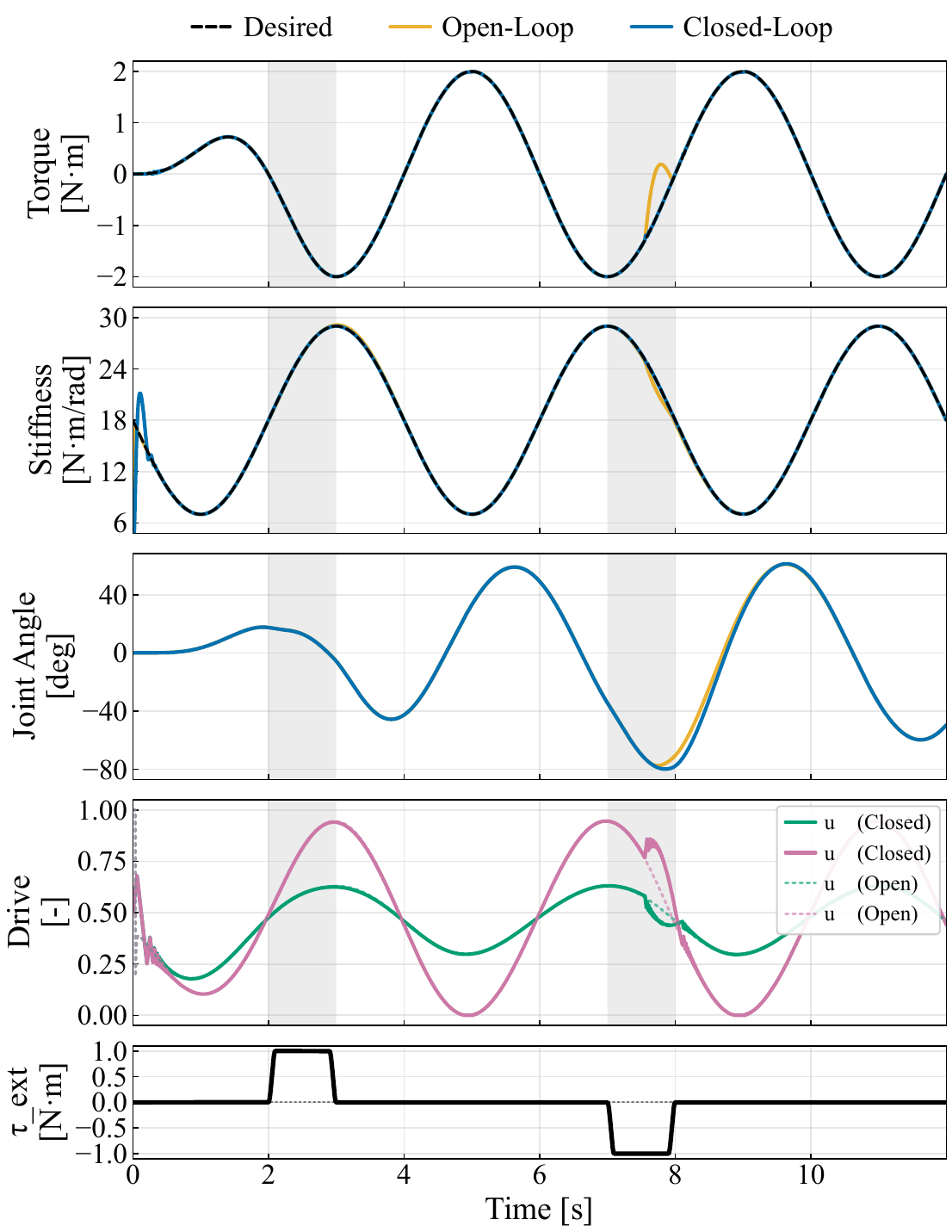}
    \caption{Response to $\pm$\SI{1.0}{\newton\meter} disturbances. Open-loop control (orange) shows disturbance-induced tracking degradation, while closed-loop control (blue) preserves independent torque and stiffness tracking with 29$\times$ lower torque error at this disturbance level. Shaded regions indicate disturbance intervals; normalized drive commands remain smooth without high-frequency switching.}
    \label{fig:controller_comparison}
\end{figure}

Validation tracks orthogonal $(T_{\text{des}}, K_{\text{mech,des}})$ trajectories under disturbances. Figure~\ref{fig:controller_comparison} shows closed-loop control maintains decoupling with 29$\times$ lower torque error than open-loop at \SI{1.0}{\newton\meter} disturbance (Table~\ref{tab:rmse_sweep}), with sub-millisecond computation (\SI{620}{\micro\second}/tick). RMSE in Table~\ref{tab:rmse_sweep} is computed over the full \SI{12}{s} run, so open-loop torque RMSE is slightly lower only at \SI{0.5}{\newton\meter}; closed-loop is markedly better at \SI{1.0}{\newton\meter} and \SI{1.5}{\newton\meter}. Closed-loop gains prioritize torque rejection, so stiffness RMSE can rise slightly due to lower stiffness-loop bandwidth.

    \begin{table}[htbp]
        \caption{Closed-loop versus open-loop tracking RMSE across disturbance sweep (full \SI{12}{s} window; superscripts O and C denote open-loop and closed-loop).}
        \label{tab:rmse_sweep}
        \centering
        \footnotesize
        \begin{tabular}{l*{4}{c}}
            \toprule
            Disturbance & $\text{RMSE}_T^{\text{O}}$ & $\text{RMSE}_T^{\text{C}}$ & $\text{RMSE}_K^{\text{O}}$ & $\text{RMSE}_K^{\text{C}}$ \\
            & [\si{\newton\meter}] & [\si{\newton\meter}] & [\si{\newton\meter\per\radian}] & [\si{\newton\meter\per\radian}] \\
            \midrule
            \SI{0.5}{\newton\meter} & \num{0.0006} & \num{0.0028} & \num{0.602} & \num{0.832} \\
            \SI{1.0}{\newton\meter} & \num{0.123} & \num{0.0042} & \num{0.627} & \num{0.832} \\
            \SI{1.5}{\newton\meter} & \num{0.288} & \num{0.097} & \num{0.711} & \num{0.872} \\
            \bottomrule
        \end{tabular}
    \end{table}

\section{Contact Interaction Tasks}\label{sec:contact}
\begin{figure}[!t]
    \centering
    \includegraphics[width=0.6\columnwidth]{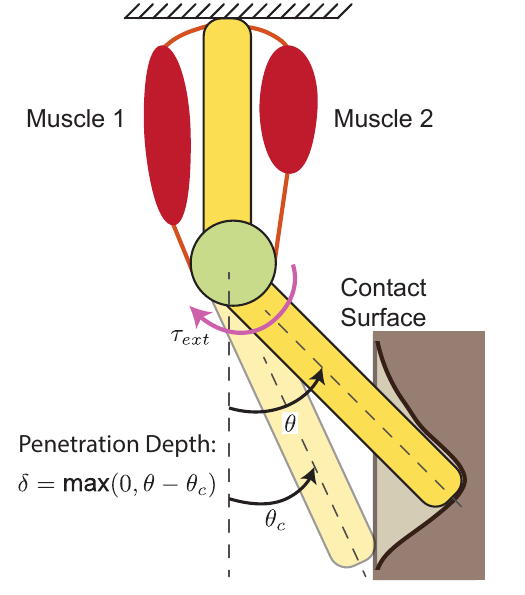}
    \caption{Antagonistic single-joint contact setup. Two opposing muscles drive the joint into a wall; contact occurs at $\theta_{\text{contact}}$, and penetration $\delta$ (in rad) generates an external reaction torque $\tau_{\text{ext}}$.}
    \label{fig:contact-setup}
\end{figure}

We evaluate depth-adaptive contact behavior without explicit surface labels (Fig.~\ref{fig:contact-setup}) using two protocols: (i) a fixed-torque stiffness-step test with direct $(T_{\text{des}},K_{\text{mech,des}})$ commands, and (ii) a coupled impedance audit where the outer loop generates $T_{\text{des}}$ and policies differ only in $K_{\text{mech,des}}$ scheduling. During contact in protocol~(ii), the outer objective is
\begin{equation}
    T_{\text{des}} = K_{\text{eff}}\!\left(\theta_{\text{ref}}-\theta\right) + B_{\text{eff}}\!\left(0-\dot{\theta}\right),
\end{equation}
with $\theta_{\text{ref}}=\SI{0.70}{rad}$; bumpless ramps keep $T_{\text{des}}$ continuous across stiffness changes. We omit explicit gravity compensation. The policy schedules $K_{\text{mech,des}}$, and unless noted otherwise we set $K_{\text{eff}}\triangleq K_{\text{mech,des}}$ after identical saturation and slew-rate limits, so stiffness scheduling influences both plant stiffness and virtual torque generation. In this section, $(K_{\text{eff}},B_{\text{eff}})$ denotes the contact-task notation corresponding to $(K_{\text{imp}},D_{\text{imp}})$ in \cref{eq:outer_loop}.

\begin{figure*}[!t]
    \centering
    \includegraphics[width=0.91\textwidth]{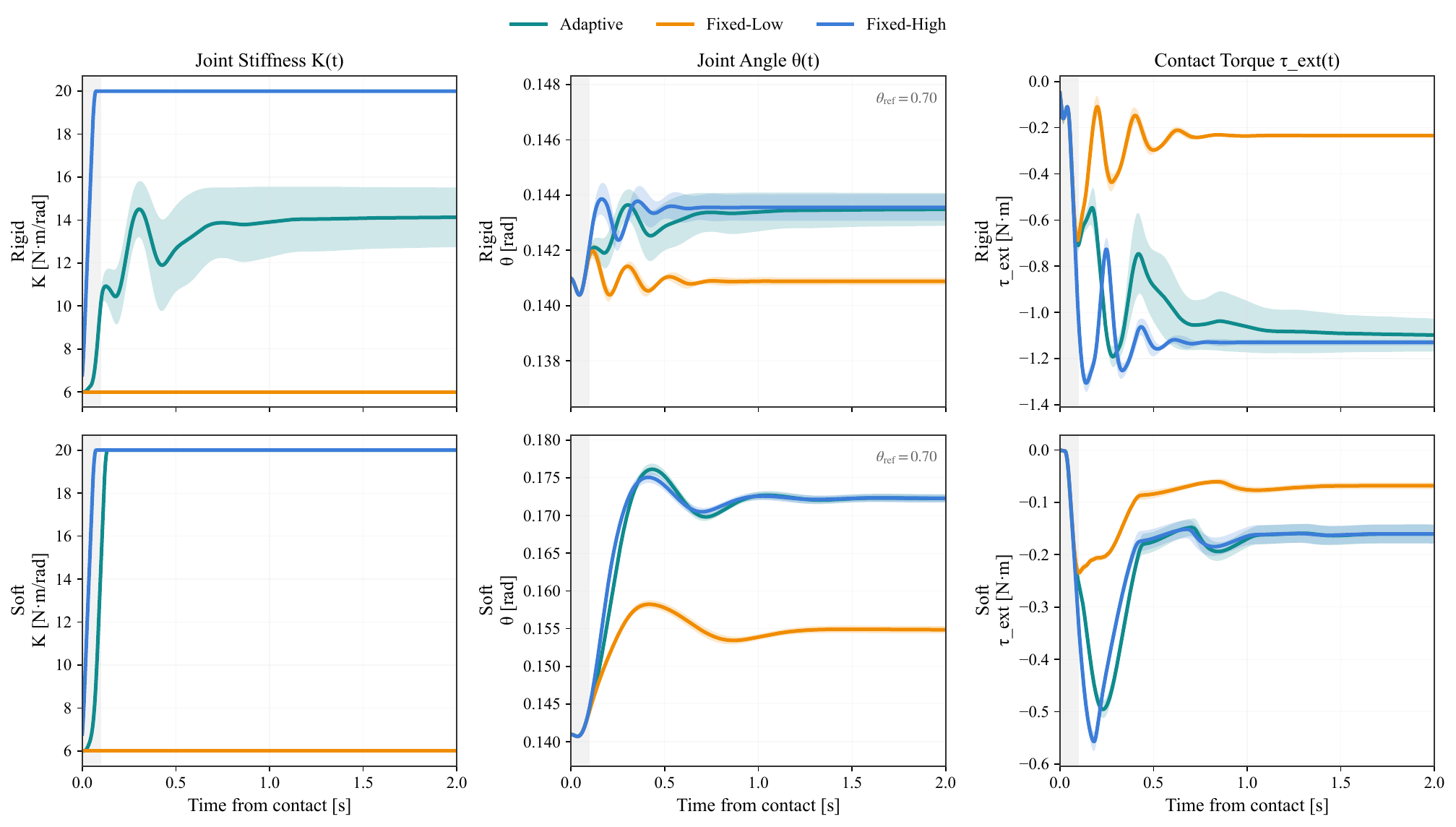}
    \caption{Contact results for depth-adaptive (teal), fixed-low (orange), and fixed-high (blue) policies (mean$\pm$std over 50 paired healthy trials). Fixed-low reduces rigid shock/load, fixed-high improves soft stability, and adaptive remains between these extremes.}
    \label{fig:contact_results}
\end{figure*}

\subsection{Policies, environments, and ablation controls}
We compare three stiffness policies:
\begin{enumerate}
    \item \textbf{Depth-adaptive (bio-inspired):} $K_{\text{mech,des}}(\delta)=K_{\text{low}}+(K_{\text{high}}-K_{\text{low}})\min\!\left(1,\delta/\delta_{\text{char}}\right)$, where $\delta=\max(0,\theta-\theta_{\text{contact}})$ is the penetration coordinate (in rad), $K_{\text{low}}=\SI{6}{\newton\meter\per\radian}$, $K_{\text{high}}=\SI{20}{\newton\meter\per\radian}$, and $\delta_{\text{char}}=\SI{0.006}{rad}$.
    \item \textbf{Fixed-low:} $K_{\text{mech,des}}=\SI{6}{\newton\meter\per\radian}$.
    \item \textbf{Fixed-high:} $K_{\text{mech,des}}=\SI{20}{\newton\meter\per\radian}$.
\end{enumerate}
All policies share the same damping gain schedule $B_{\text{eff}}(t)$: $B_{\text{impact}}$ for \SI{20}{ms}, a linear ramp to $B_{\text{stance}}$ over \SI{80}{ms}, then $B_{\text{stance}}$. In the reported runs, $B_{\text{impact}}=B_{\text{stance}}=\SI{0.2}{\newton\meter\second\per\radian}$, so $B_{\text{eff}}$ is constant apart from the shared per-trial multiplier.

Contact uses spring-damper environments in three stiffness regimes: soft (\SI{5}{\newton\meter\per\radian}), medium (\SI{60}{\newton\meter\per\radian}), and rigid (\SI{500}{\newton\meter\per\radian}), with threshold $\theta_{\text{contact}}=\SI{0.14}{rad}$ and penetration $\delta=\max(0,\theta-\theta_{\text{contact}})$. For $\delta>0$, $\tau_{\text{ext}}=-s\,c\,K_{\text{env}}\,\tilde{\delta}-s\,c_{\text{eff}}\,\tilde{\delta}^{p}\,[\dot{\theta}]_{+}\,g_{\delta}$, with soft contact activation $c=\frac{1}{2}(1+\tanh((\theta-\theta_{\text{contact}})/\epsilon_c))$, $\epsilon_c=\SI{0.012}{rad}$, $[\dot{\theta}]_{+}=\max(0,\dot{\theta})$, $c_{\text{eff}}=2\zeta_{\text{env}}\sqrt{J_{\mathrm{eq}}K_{\text{env}}}+c_{\text{base}}$, $c_{\text{base}}=\SI{2.0}{\newton\meter\second\per\radian}$, and $p\ge0$ a penetration exponent shaping depth-dependent damping ($p=0$ in the implemented Kelvin--Voigt model). Here $\tilde{\delta}$ is the smoothed unilateral depth, $g_{\delta}$ smoothly suppresses damping near detach, and $s$ is a short soft-start factor over \SI{4}{ms}.

Trials start from light compression at $\theta_0=\SI{0.141}{rad}$ and $\dot{\theta}_0=\SI{0.0}{rad/s}$. The coupled audit uses 50 paired healthy trials per surface (\SI{5}{s}, controller and integration both at \SI{1}{ms}). The stiffness-step test uses 10 paired trials per condition, and the strict health-gated 2$\times$2 audit uses 50 paired trials per quadrant/profile.

All policies use identical initialization, limits, and paired perturbations (shared within each trial) on environment stiffness, stance damping, and contact damping ratio.

\paragraph{Trial health gating}
A trial is labeled \emph{healthy} if contact remains physically plausible and numerically stable over the evaluation window (no divergence, no solver failure, no sustained detachment/chattering, and no persistent command-limit pathologies). Paired comparisons retain only perturbation draws that are healthy for all compared policies.

\paragraph{Ablation runs (attribution controls)}
Because $K_{\text{eff}}\equiv K_{\text{mech,des}}$, outcomes in the main protocol can reflect both outer-gain effects and mechanical-stiffness effects. We therefore run a 2$\times$2 attribution audit comparing fixed-low (FL) versus fixed-high (FH) under two switches: factor A toggles whether outer stiffness gain $K_{\text{eff}}$ differs between FL and FH or is held constant, and factor B toggles whether mechanical stiffness target $K_{\text{mech,des}}$ differs between FL and FH or is held constant. This yields four quadrants: A0B0 (both held constant), A1B0 (virtual-only), A0B1 (mechanical-only), and A1B1 (coupled). When held constant, the outer gain and mechanical stiffness are set to \SI{12}{\newton\meter\per\radian} (with the same saturation and slew-rate limits as varying cases). We analyze these controls in the Results subsection.

\subsection{Results}
For the coupled audit (Fig.~\ref{fig:contact_results} and Tables~\ref{tab:contact-stats}--\ref{tab:contact-metrics}), we report impact shock peak torque $\max_{t\in[0,100\text{ms}]} |\tau_{\text{ext}}(t)|$, sustained cumulative load $L_{\tau}\triangleq\int_{0.3}^{5.0} |\tau_{\text{ext}}(t)|\,dt$, and a stability metric (fraction of time after \SI{0.3}{s} satisfying $|\theta(t)-\theta_{\mathrm{ss}}|\le 0.005|\theta_{\mathrm{ss}}|$), where $t=0$ is the first-contact instant after contact-time alignment. We define $\theta_{\mathrm{ss}}$ per trial as the median of $\theta(t)$ over the last \SI{0.3}{s} of the post-contact trajectory. We include both shock peak and $L_{\tau}$ to distinguish transient impact severity from sustained contact exposure. Metrics are computed per trial and reported as mean $\pm$ std; pairwise claim comparisons use two-sided Wilcoxon signed-rank tests with Holm correction across the three pairwise contrasts per claim, except rigid-shock FL-Bio, which is treated as a practical closeness check ($|\Delta|\le\SI{0.05}{\newton\meter}$), while the directional fixed-low (FL) versus fixed-high (FH) audit in Table~\ref{tab:factorial-causal} uses one-sided tests.

\paragraph{Primary decoupling capability (fixed-torque stiffness-step)}
Table~\ref{tab:capability-summary} reports the fixed-torque stiffness-step test in contact. Across 12 paired conditions (4 torque targets $\times$ 3 profiles), the decoupled controller reduces torque disturbance during stiffness transitions versus the stiffness-coupled baseline.
In Table~\ref{tab:capability-summary}, shock peak is evaluated over the first \SI{200}{ms} (capability-specific window).
\begin{table}[t]
  \caption{Fixed-torque stiffness-step capability summary\\12 paired conditions (3 profiles $\times$ 4 targets; 10 trials/condition).}
  \label{tab:capability-summary}
  \centering
  \scriptsize
  \setlength{\tabcolsep}{3pt}
  \begin{tabular}{lccc}
    \toprule
    Metric & Dec better (n/12) & Sig. ($p<0.05$) & Worst-case $\Delta$ \\
    \midrule
    Torque bump & 12/12 & 12/12 & $-2.23$~\si{\newton\meter} \\
    Cum. load $L_{\tau}$ & 12/12 & 12/12 & $-0.21$~\si{\newton\meter\second} \\
    SS torque error & 12/12 & 12/12 & $-1.43$~\si{\newton\meter} \\
    $|\Delta T_{\mathrm{des}}|$ & 12/12 & 12/12 & $-2.31$~\si{\newton\meter} \\
    Shock peak & 12/12 & 11/12 & $-0.44$~\si{\newton\meter} \\
    \bottomrule
  \end{tabular}
  \par\vspace{0.55ex}
  {\raggedright\footnotesize\emph{Notes:} Worst-case $\Delta$ is the least favorable condition-level mean difference (Dec$-$Coup; negative favors decoupled). Torque bump = peak $|T-T_{\mathrm{pre}}|$ during the step; SS torque error = mean $|T-T_{\mathrm{des}}|$ in the final steady window; $|\Delta T_{\mathrm{des}}|$ = peak desired-torque change from the coupled baseline during the step; shock peak (step) = max $|\tau_{\mathrm{ext}}|$ in the first \SI{200}{ms}.\par}
\end{table}

\paragraph{Coupled contact outcomes (depth-adaptive vs fixed stiffness)}
Figure~\ref{fig:contact_results} and Tables~\ref{tab:contact-stats}--\ref{tab:contact-metrics} summarize the $n=50$ paired healthy audit. Fixed-low and fixed-high recover the expected rigid shock/load versus soft-stability tradeoff, and depth-adaptive remains between them without surface labels.
\begin{table}[t]
  \caption{Primary contact claim checks (paired two-sided Wilcoxon with Holm-adjusted $p$-values, $n=50$).}
  \label{tab:contact-stats}
  \centering
  \begingroup
  \setlength{\tabcolsep}{1.4pt}
  \renewcommand{\arraystretch}{1.08}
  \footnotesize
  \begin{tabular}{@{}lccc@{}}
    \toprule
    \textbf{Claim (order)} & \textbf{Means} & \textbf{$\Delta$ pairs} & \textbf{Holm-adj. $p$} \\
    & \textbf{(FL/Bio/FH)} & \textbf{(FL-Bio/Bio-FH/FL-FH)} & \textbf{(3 pairs)} \\
    \midrule
    \shortstack[l]{Rigid shock\\(FL$\approx$Bio$<$FH)} & \shortstack[c]{0.71/0.72/\\1.04} & \shortstack[c]{-0.019/-0.319/\\-0.338} & \shortstack[c]{--/\textbf{$<10^{-4}$}/\\\textbf{$<10^{-4}$}} \\
    \shortstack[l]{Rigid load $L_{\tau}$\\(FL$<$Bio$<$FH)} & \shortstack[c]{1.11/5.09/\\5.32} & \shortstack[c]{-3.98/-0.23/\\-4.21} & \shortstack[c]{\textbf{$<10^{-4}$}/\textbf{$<10^{-4}$}/\\\textbf{$<10^{-4}$}} \\
    \shortstack[l]{Soft stability\\(FL$<$Bio$<$FH)} & \shortstack[c]{86.46/89.62/\\91.02} & \shortstack[c]{-3.16/-1.40/\\-4.55} & \shortstack[c]{\textbf{$<10^{-4}$}/\textbf{$<10^{-4}$}/\\\textbf{$<10^{-4}$}} \\
    \bottomrule
  \end{tabular}
  \par\vspace{0.95ex}
  {\raggedright\footnotesize\emph{Notes:} Units: shock [\si{\newton\meter}], cumulative load $L_{\tau}$ [\si{\newton\meter\second}], stability mean [\%] with differences in percentage points. FL=fixed-low, Bio=adaptive, FH=fixed-high. Means are descriptive; $\Delta$ and Holm-adjusted $p$-values are inferential (pair order: FL-Bio/Bio-FH/FL-FH), using two-sided paired Wilcoxon tests with Holm correction across the three pairwise comparisons per claim. For rigid shock, FL-Bio is a closeness check ($|\Delta_{\mathrm{FL-Bio}}|=0.019$~\si{\newton\meter}, threshold $\le 0.05$~\si{\newton\meter}), so adjusted $p$ is reported only for Bio-FH and FL-FH.\par}
  \endgroup
\end{table}

\begin{table}[t]
  \caption{Contact metrics across soft, medium, and rigid regimes\\(50 paired trials; mean $\pm$ std).}
  \label{tab:contact-metrics}
  \centering
  \begingroup
  \setlength{\tabcolsep}{2.5pt}
  \scriptsize
  \resizebox{\columnwidth}{!}{%
  \begin{tabular}{@{}llcccc@{}}
    \toprule
    \textbf{Surf.} & \textbf{Pol.} & \textbf{Shock} & \textbf{Load $L_{\tau}$} & \textbf{Stab.} & \boldmath$\theta_{\text{ss}}$ \\
     &  & [\si{\newton\meter}] & [\si{\newton\meter\second}] & [\%] & [rad] \\
    \midrule
    Soft   & Bio & 0.25$\pm$0.01 & 0.78$\pm$0.08 & 89.62$\pm$0.28 & 0.172$\pm$0.001 \\
           & FL  & \textbf{0.24$\pm$0.01} & \textbf{0.33$\pm$0.03} & 86.46$\pm$0.57 & 0.155$\pm$0.000 \\
           & FH  & 0.33$\pm$0.01 & 0.77$\pm$0.08 & \textbf{91.02$\pm$0.56} & 0.172$\pm$0.001 \\
    \midrule
    Medium & Bio & 0.43$\pm$0.02 & 3.71$\pm$0.18 & 97.13$\pm$1.29 & 0.155$\pm$0.001 \\
           & FL  & \textbf{0.41$\pm$0.02} & \textbf{0.91$\pm$0.02} & 95.40$\pm$0.58 & 0.145$\pm$0.000 \\
           & FH  & 0.58$\pm$0.04 & 3.70$\pm$0.18 & \textbf{98.62$\pm$1.02} & 0.155$\pm$0.001 \\
    \midrule
    Rigid  & Bio & 0.72$\pm$0.02 & 5.09$\pm$0.34 & 97.28$\pm$2.02 & 0.143$\pm$0.001 \\
           & FL  & \textbf{0.71$\pm$0.02} & \textbf{1.11$\pm$0.01} & \textbf{99.99$\pm$0.07} & 0.141$\pm$0.000 \\
           & FH  & 1.04$\pm$0.04 & 5.32$\pm$0.06 & 99.98$\pm$0.06 & 0.144$\pm$0.000 \\
    \bottomrule
  \end{tabular}%
  }
  \par\vspace{0.55ex}
  {\raggedright\footnotesize\emph{Notes:} Shock = max $|\tau_{\text{ext}}|$ in first \SI{100}{ms}. Cumulative load $L_{\tau}$ (\(L_1\) norm) $=\int_{0.3}^{5.0}|\tau_{\text{ext}}|dt$ (time-integrated absolute contact torque). Stab. = \% time in a 0.5\% band after \SI{0.3}{s}. Bold marks surface-wise best values for Shock, Load, and Stab. (Shock/Load lower is better; Stab. higher is better). $\theta_{\mathrm{ss}}$ is reported for context (steady-state joint angle) and is not ranked.\par}
  \endgroup
\end{table}

\paragraph{Ablation outcomes (attribution controls)}
The compact 2$\times$2 audit (Table~\ref{tab:factorial-causal}) attributes rigid peak/load and soft-stability effects under strict health gating (all cells: 100\% healthy trials). In nominal trials, both virtual scheduling (A1B0) and mechanical scheduling (A0B1) contribute, with the largest rigid reduction in A1B1. In challenge trials ($\dot{\theta}_0=\SI{0.2}{rad/s}$), rigid load reductions persist, but soft-stability attribution becomes non-additive (A1B1 flips sign), indicating profile-sensitive interaction.
\begin{table}[t]
  \caption{Compact 2$\times$2 attribution audit (rigid peak/load + soft stability).}
  \label{tab:factorial-causal}
  \centering
  \begingroup
  \setlength{\tabcolsep}{2.7pt}
  \footnotesize
  \begin{tabular}{@{}llccc@{}}
    \toprule
    Profile & Quadrant & Rig. Peak $\Delta$ & Rig. Load $\Delta$ & Soft Stab. $\Delta$ \\
    \midrule
    N & A0B0 (both fixed) & $+0.00$ & $+0.00$ & $+0.00$ \\
    N & A1B0 (virtual only) & $-0.10^{*}$ & $-2.95^{*}$ & $+3.76^{*}$ \\
    N & A0B1 (mechanical only) & $-0.17^{*}$ & $-2.36^{*}$ & $+8.41^{*}$ \\
    N & A1B1 (coupled) & $-0.34^{*}$ & $-4.20^{*}$ & $+4.57^{*}$ \\
    \midrule
    C & A0B0 (both fixed) & $+0.00$ & $+0.00$ & $+0.00$ \\
    C & A1B0 (virtual only) & $+0.00^{*}$ & $-2.94^{*}$ & $+2.94^{*}$ \\
    C & A0B1 (mechanical only) & $-0.04^{*}$ & $-2.36^{*}$ & $+6.65^{*}$ \\
    C & A1B1 (coupled) & $-0.02^{*}$ & $-4.18^{*}$ & $-4.94$ \\
    \bottomrule
  \end{tabular}
  \par\vspace{0.35ex}
  {\raggedright\footnotesize\emph{Notes:} N=nominal, C=challenge ($\dot{\theta}_0=\SI{0.2}{rad/s}$). Deltas are FL--FH for rigid metrics and FH--FL for soft stability. $^{*}$ denotes one-sided paired Wilcoxon $p<0.05$ (50 paired trials/cell). Values are rounded for compactness; significance is computed on unrounded paired differences.\par}
  \endgroup
\end{table}

\paragraph{Mechanism and intuition}
Because $K_{\text{eff}}\equiv K_{\text{mech,des}}$, lower stiffness reduces restoring torque and can induce sinking on compliant surfaces. Depth-adaptive scheduling increases stiffness with penetration, remaining near $K_{\text{low}}$ at small penetrations (typical of stiff contacts) and ramping toward $K_{\text{high}}$ as penetration grows on compliant contacts, supporting shock--stability adaptation without explicit surface classification \cite{Zajac1989,houk1981regulation}.

\section{Conclusion}
We presented a unified framework for antagonistic artificial muscles that combines a separable Padé [2/1] force law, a minimal two-state dynamic wrapper, and a cascaded controller in co-contraction/bias coordinates. The identified HASEL model achieves sub-Newton force prediction error and supports sub-millisecond torque-stiffness control in simulation.

Contact experiments show two complementary results: (i) fixed-torque stiffness-step tracking preserves torque through stiffness transitions across nominal, approach-velocity, and delayed/noisy profiles; (ii) in the coupled protocol, fixed-low/fixed-high expose a rigid shock/load versus soft-stability tradeoff, while depth-adaptive scheduling remains between these fixed extremes without explicit surface labels. In the compact 2$\times$2 audit, nominal-profile rigid outcomes reflect both virtual and mechanical scheduling, while challenge-profile outcomes show stronger profile dependence.

\paragraph{Limitations and future work}
Key limitations remain before deployment: contact results are simulation-based in simplified linear spring-damper environments, identification is HASEL-only, and the current study assumes a symmetric single-DOF antagonistic pair. Next steps are hardware replication on HASEL/PAM/DEA platforms and extension to coupled multi-DOF systems, including controlled comparison with learning-based policies under the same paired-perturbation protocol.

Ultimately, this framework provides a practical path to musculoskeletal robots that use torque--stiffness control as a first-class interaction primitive for safer adaptive behavior in unstructured real-world environments.

\bibliographystyle{IEEEtran}
\bibliography{refs}

\end{document}